\documentclass[preprint,review,12pt,3p,sort&compress]{elsarticle}

\usepackage{tabularx}
\usepackage{subfig}
\usepackage{placeins}
\usepackage{changes}
\usepackage{footnote}
\makesavenoteenv{tabularx}
\makesavenoteenv{table}
\usepackage[linesnumbered,lined,boxed,commentsnumbered]{algorithm2e}
\usepackage{amsmath}

\begin{document}
\begin{frontmatter}
\title{An evolutionary approach to continuously estimate CPR quality parameters from a wrist-worn inertial sensor}

\author[label1,label2]{Christian Lins\corref{cor1}} 
\address[label1]{Carl von Ossietzky University Oldenburg, Assistance Systems and Medical Device Technology, Dep. Health Services Research, Ammerl\"ander Heerstr. 140, 26129 Oldenburg, Germany}
\address[label2]{OFFIS -- Institute for Information Technology, Div. Health, Escherweg 2, 26121 Oldenburg, Germany\fnref{label4}}

\cortext[cor1]{Corresponding author}

\ead{christian.lins@offis.de}

\author[label1]{Bj\"{o}rn Friedrich}

\author[label1,label2]{Andreas Hein}

\author[label1]{Sebastian Fudickar}

\begin{abstract}

Cardiopulmonary resuscitation (CPR) is one of the most critical emergency interventions for sudden cardiac arrest. In this paper, a robust sinusoidal model-fitting method based on a Evolution Strategy inspired algorithm for CPR quality parameters -- naming chest compression frequency and depth -- as measured by an inertial measurement unit (IMU) attached to the wrist is presented. The proposed approach will allow bystanders to improve CPR as part of a continuous closed-loop support system once integrated into a smartphone or smartwatch application. By evaluating the model's precision with data recorded by a training mannequin as reference standard, a variance for the compression frequency of $\pm 2.22$ compressions per minute (cpm) has been found for the IMU attached to the wrist. It was found that this previously unconsidered position and thus, the use of smartwatches is a suitable alternative to the typical placement of phones in hand for CPR training. 
\end{abstract}


\end{frontmatter}

\section{Introduction}

Sudden cardiac arrest (SCA) is one of the most prominent diseases (350,000-700,000 individuals a year in Europe are affected \cite{Berdowski2010, Grasner2011, Grasner2013}). SCA can significantly affect the independent living of the victims if medical treatment is not available within a few minutes \cite{Perkins2015,DeMaio2003}. In the case of cardiac arrest, the transport of oxygen and glucose to the human body's cells stops immediately due to the disrupted heart function. This leads to irreparable cell damage if blood circulation is not quickly reestablished, e.g. supported by cardiopulmonary resuscitation (CPR). If the blood circulation is interrupted, the cells of the nervous system, including the brain, reduce their functionality after 10 seconds, which leads, for example, to unconsciousness \cite{Schmidt2011}. The death of the cells begins after about 3 minutes without blood circulation~\cite{Schmidt2011}.

Medical personnel such as paramedics are trained in Advanced Life Support (ALS)~\cite{Soar2015} methodology that includes CPR. Unfortunately, paramedics are usually not immediately available if a cardiac arrest occurs in the field. The typical median reaction time of paramedics is about 5-8 minutes \cite{Perkins2015} and with every minute the chances of survival of the victims without CPR decrease. Thus, victims rely on the initial CPR support of bystanders within the first so-called \emph{golden minutes} after cardiac arrest to prevent long-term adverse effects or death. Since these bystanders can offer essential initial resuscitation support, corresponding technical solutions to support them with online feedback regarding the quality of CPR are beneficial. 

\begin{figure}[b]
    \centering
    \includegraphics[width=\textwidth]{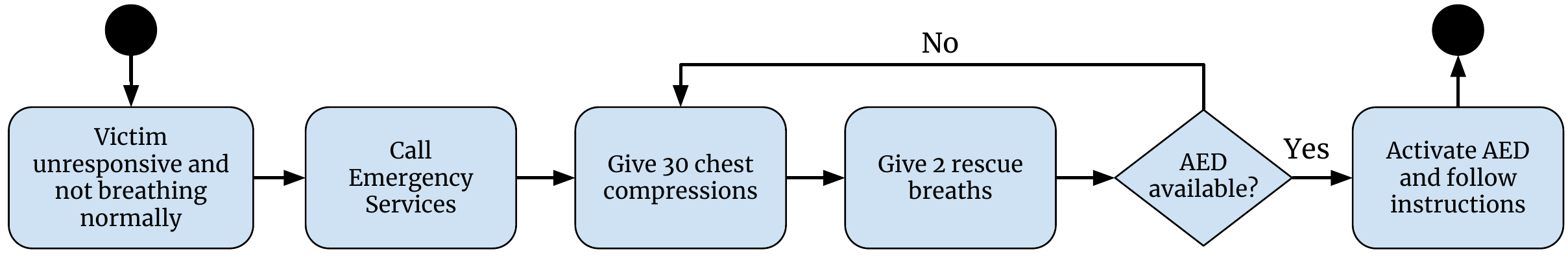}
    \caption{Basic Life Support Algorithm. AED is automated external defibrillator.}
    \label{fig:blsalgo}
\end{figure}

For the following discussion, the optimal Basic Life Support (BLS) \cite{Perkins2015} procedure (the methodology aiming for non-specialists) is worthwhile to be briefly recapitulated (see \mbox{Figure \ref{fig:blsalgo}}): In case of a cardiac arrest, it is essential to ensure sufficient oxygenation of the nerve cells via correctly conducted cardiac massages (chest compressions) as the most critical countermeasure. During this cardiac massage, the heart is compressed by orthogonal pressure onto the breastbone. In order to sustain a minimal blood circulation, a chest compression frequency (CCF) of the cardiac massage should range within 100-120 \emph{compressions per minute} (cpm), and a chest compression depth (CCD) of 5-6 cm is required. With a CCF below 100 cpm, blood circulation is insufficient to fullfill the essential tissue oxigenation, while with a CCF over 120 cpm the heart complete refillment with blood is not assured, thus resulting in to little amount of blood being circulated in the next compression. Ideally, but not necessarily, the procedure is combined with rescue breathing, to improve the chance of survival and reduce neurological deficits \cite{Perkins2015}.

While typical bystanders can develop a sufficient feeling of semi-ideal compression depth, the consistent application of the correct compression frequency and depth is challenging, especially for extended periods of cardiac massage with the associated muscle-fatigue and mental pressure. Thus, instant feedback regarding the correct execution (regarding CCD and CCF) during a cardiac arrest will be beneficial for untrained bystanders. Such feedback could be derived from online monitoring of the quality of the cardiac massage from the vertical acceleration measured by inertial measurement units (IMUs) and giving continuous feedback and adjustment hints. Corresponding smartphone applications are well suited for this purpose due to their high availability \cite{Kalz2014, Ahn2016}. Renshaw et al. have confirmed the general benefit regarding CCF/CCD for the BHF PocketCPR application and recognized an improved performance (from 66 to 91 cpm) and increased confidence of bystanders \cite{Renshaw2017}. However, for such CPR training apps, accurate CPR information (regarding CCF and CCD) is an essential requirement \cite{Ahn2016}, which is achieved partially by the existing implementations (as summarized in Table \ref{tab:cpr_training_ref}).

\begin{table}[ht!]
    \caption{Accuracies of existing on-body chest-compression algorithms.}
    \label{tab:cpr_training_ref}
    \centering
    \small
    \begin{tabularx}{\linewidth}{p{1.9cm}|p{4.9cm}|p{1.5cm}|p{1.5cm}|p{2.1cm}|p{1.6cm}}
        Position & Algorithm / System & CCF Error [cpm] & CCD Error
[mm] & Reference System & Year\\\hline

On chest &	Spectral techniques on short acceleration intervals (FFT-based) &	$<$ 1.5 &	$<$ 2 &	photoelectric distance sensor &	2014 \cite{Gonzalez-Otero2014}\\

On chest &	Butterworth HP filter, 2x integration, manual reset	& - &	1.6 (within 95\%) &
	mannequin potentiometer &
	2002 \cite{Aase2002}\\

On chest &	Weighted smoothing, double 2x (transient component emphasizing + integration), peak detection (U-CPR)&
	- &	1.43 mean (SD 1.04) &	mannequin potentiometer &
	2015 \cite{Song2015}\\

On chest&	PocketCPR, method unknown&	- &	1.01 mean  (SD 0.74) &	mannequin potentiometer &
	2015 \cite{Song2015}\\

On chest&	Spectral analysis of acceleration&	0.9 median&	1.3 median &	displacement sensor & 2016 \cite{DeGauna2016} 
\\
Wrist & SamCPR, method unknown & - & - & - & 2015 \cite{samcpr} \\
Wrist & Smart-watch life saver App & - & - & - & 2015 \cite{Gruenerbl2015}\\
    \end{tabularx}
\end{table}

Most of these approaches use Fast Fourier Transformation (FFT) to determine the regular frequencies from the inertial data and identify the frequency spectrum's central frequencies via peak detection. While being straightforward, this approach is susceptible to erroneous peak selection and a resulting frequency shift of the CCFs by orders of magnitudes. 
Also, the double-integration of the acceleration signal -- a standard processing step to determine the displacement-vector as a preprocessing step for the CCD -- is challenging due to the signal drift if the accelerometer is not perfectly aligned with the gravity axis \cite{ladetto2000foot}.
Consequently, the typical integration process is inherently unstable and leads to relevant errors unless boundary conditions are applied for each compression cycle (e.g. in very short windows)~\cite{DeGauna2016}. 

\begin{figure}[h]
    \centering
    \includegraphics[width=0.9\textwidth]{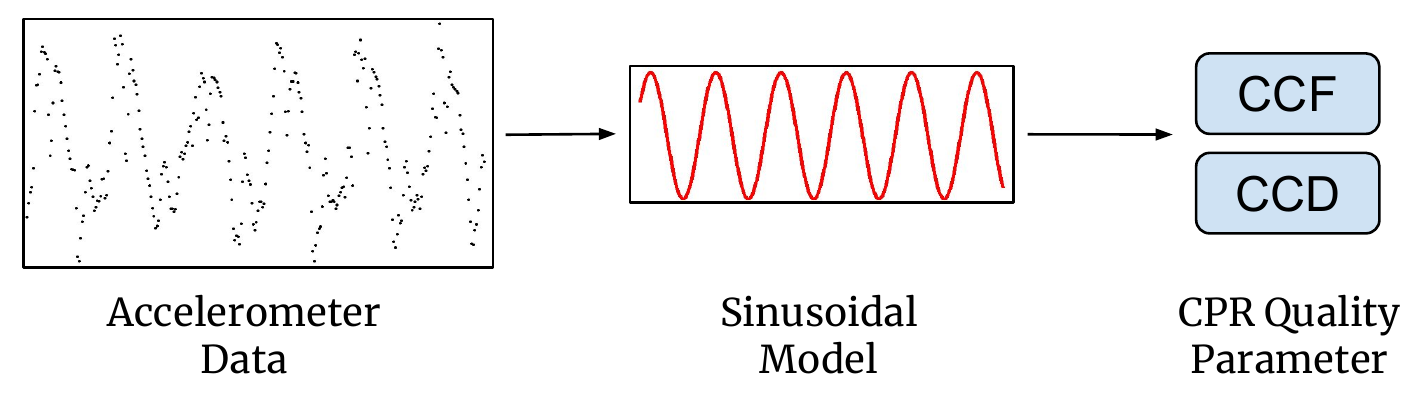}
    \caption{Basic approach of a sinusoidal model for CPR quality parameter estimation (see Section \ref{sec:system_approach} and Figure \ref{fig:system_concept} for more details). The accelerometer data is measured by an IMU sensor and then window-based fitted to a sine function, which in turn provides the CCF/CCD parameters sought.}
    \label{fig:cprapproach}
\end{figure}

The motion in cardiopulmonary resuscitation is primarily a rhythmic movement that approximates the natural rhythm of the human heartbeat.
Thus, the use of a robust sinusoidal model, i.e., sine curve, might not suffer from disadvantages of the FFT and its subsequent effects on the derived CPR parameters (see Figure \ref{fig:cprapproach}), due to its implicit periodic accordance with the CPR. A concept that was shown for depth-image based motion capture of CPR movements \cite{Lins2018, Lins2019}, but still needs to be confirmed for use with acceleration data recorded directly via IMUs at rescuers.  

Furthermore, the discussed algorithms have been mainly evaluated for the use of IMUs in a grasp-in-hand use (i.e., the sensor or smartphone is held between the back of the lower hand and the palm of the upper hand during the procedure). A position that has been reported to be a slightly uncomfortable and disturbing positioning for bystanders \cite{Park2016}, which as well might mislead them into learning incorrect postures \cite{Park2017}. To overcome this drawback, Park et al. \cite{Park2016} proposed the fixation of smartphones via an armband on the dorsum manus or at the arm and had shown an increased convenience compared to the common grasp-in-hand approach. However, they reported a reduced sensitivity, which they explained with the amplified inertial forces resulting from the additional device's swing. Similarly, Ruiz de Gauna et al. \cite{Gauna2015} compared sensor placement at the dorsum manus with one fixed to the forearm 7 cm above the wrist. They confirmed a significantly increased error for the forearm placement.

Consequently, while IMUs have generally proven to be a precise and practical approach to measuring chest compression depth and frequency during CPR, user-friendliness via smartphones is a challenging task as it affects the quality of CPR for bystanders. Smartwatches hold benefits over the use of smartphones regarding usability. They could be expected to achieve higher reliability towards arm-movements because they are potentially less affected by hand movements and rigidly attached to the wrist. Furthermore, they overcome challenging aspects of reduced tactile pressure sensation at the hands. However, the use of IMUs on alternative placements was repetitively found challenging for sufficient accuracy. The suitability of smartwatches for CPR training and online-support regarding the sensitivity of CCD and CCF detection has yet to be investigated. 

Our approach is the combination of a wrist-worn IMU with a sinusoidal model. To fit the accelerometer data to the sine curve of the model, an appropriate optimization algorithm is required. Evolutionary methods have already been found to be principally suitable for a similar approach: Lins et al. \cite{Lins2019} use a Differential Evolution (DE) algorithm to fit optical motion capture data to a sinusoidal model. In this context DE is used for continuous window-based curve fitting of the model by optimizing the parameters of a sine function. The article at hand proposes a Evolution Strategy (ES) inspired algorithm for the continuous model-fitting (i.e., without complete reinitialization of the population) with IMU data. 

Nature-inspired algorithms such as a Genetic Algorithm (GA) have also already proven useful in signal processing \cite{Upadhyay2014, Kumar2015, Aggarwal2015}. In addition, there are already some work that has used GAs for curve fitting, e.g. Zhao et al. use a GA to fit the parameters of a Bezier curve \cite{Zhao2013}.  GAs have also been used to fit polynomials \cite{Clegg2005}. In the recent work of Jiang et al. a modified GA is used to optimize the parameters of sine signals \cite{Jiang2019,Jiang2020}.
Consequently, in this study, we aim to investigate the following two research questions: 

\begin{enumerate}
    \item How suitable are wrist-worn inertial sensors (e.g. smartwatches) for the online detection of the chest compression during CPR regarding the accuracy of resulting CCF and CCD parameters compared to both the Resusci Anne training mannequin and the typical hand-holding as the gold standard?
    
    \item How suitable is the algorithmic approach for fitting a sinusoidal model of the chest compression during CPR regarding the accuracy of resulting CCF and CCD parameters for the considered sensor placements compared to the Resusci Anne training mannequin as reference system?
\end{enumerate}

In Section 2, our system approach, details of the used algorithm,  study design, and applied evaluation methodology is introduced. In Section 3, the results of the study are presented and discussed in Section 4. The article is concluded in Section 5. 

\section{Materials and Methods}

\subsection{System Approach}
\label{sec:system_approach}

\begin{figure}[h]
    \centering
    \includegraphics[width=0.9\textwidth]{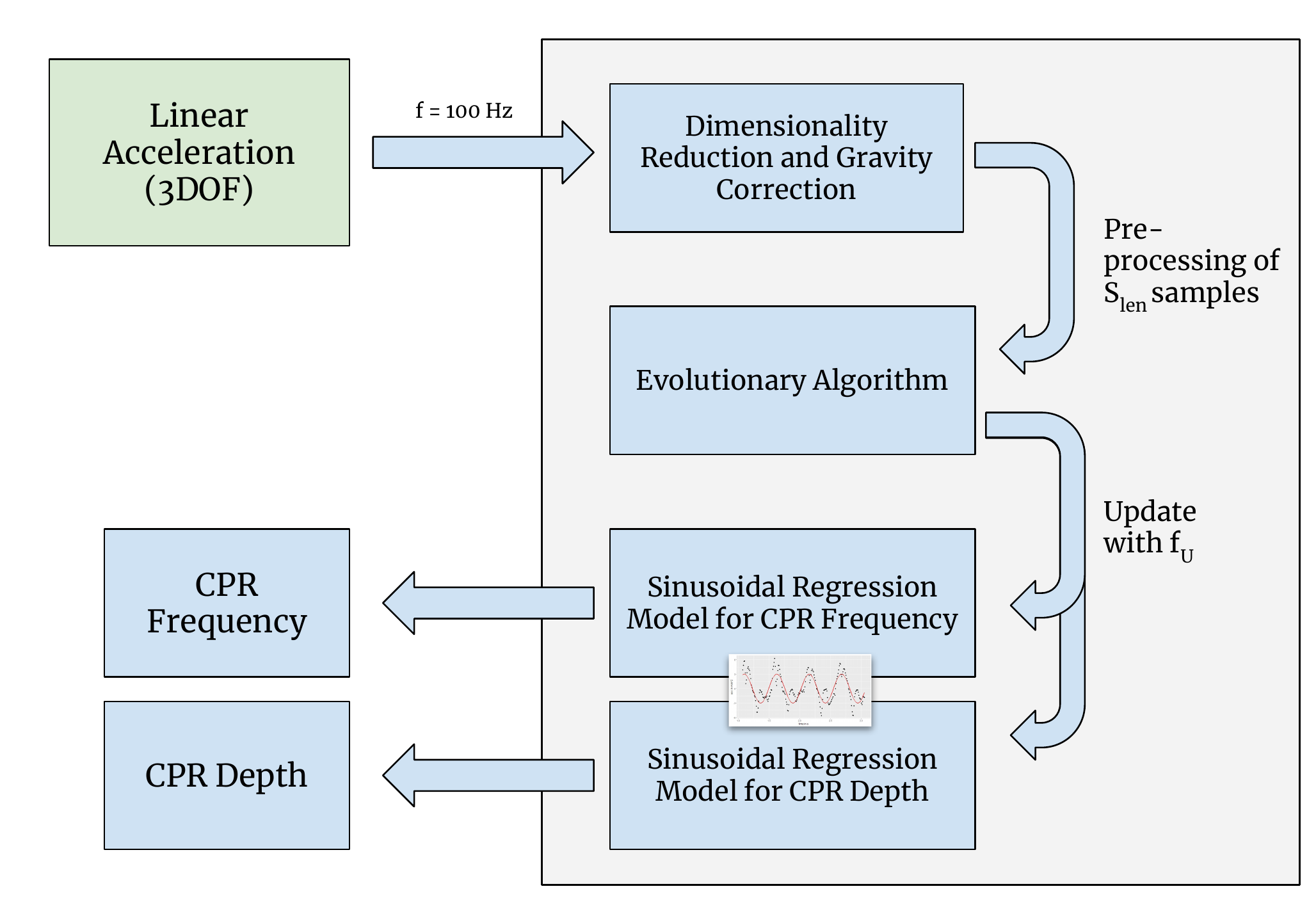}
    \caption{System concept overview of the proposed method.}
    \label{fig:system_concept}
\end{figure}

In the proposed system, the curve of chest compressions during CPR is (Figure \ref{fig:system_concept}) analyzed from data of a wrist-worn (and hand-held as reference) three-dimensional inertial sensor containing accelerometer, gyroscope, and magnetometer (9DOF). 
The accelerometer signal contains earth gravity as a constant component distributed among all three accelerometer axis, depending on the sensor orientation. 
For fitting the model function, only the acceleration caused by the orthogonal pressure on the patient's chest is relevant. 
Unfortunately, when placed on the wrist, the orientation of the sensor is never entirely orthogonal to the chest. So the earth's gravity is distributed over all three axes with different magnitudes.
The approach to handle this is to consider only the total acceleration $a$, i.e. the Euclidean distance over all axes and subtract the constant gravity ($9.81 \frac{m}{s^2}$) from the total acceleration, which is one-dimensional (see Equation \ref{eq:linear_acc}).

\begin{equation}
    a_i=\sqrt{\left({a_{i,x}}^2+{a_{i,y}}^2+{a_{i,z}}^2\right)} - 9.81 \frac{m}{s^2}, i \in \{1,\dots,S_{len}\}
    \label{eq:linear_acc}
\end{equation}


This one-dimensional time-series data are window-based (window length $S_{len}$) fitted to a sinusoidal function model. The fitting generates adapted models per ${f_U}^{-1}$ seconds, from which the CPR parameters frequency (CCF) and compression depth (CCD) can be derived (recap Figure \ref{fig:system_concept}).

\subsection{Model Function}

\begin{figure}[htb]
    \centering
    \includegraphics[width=\textwidth]{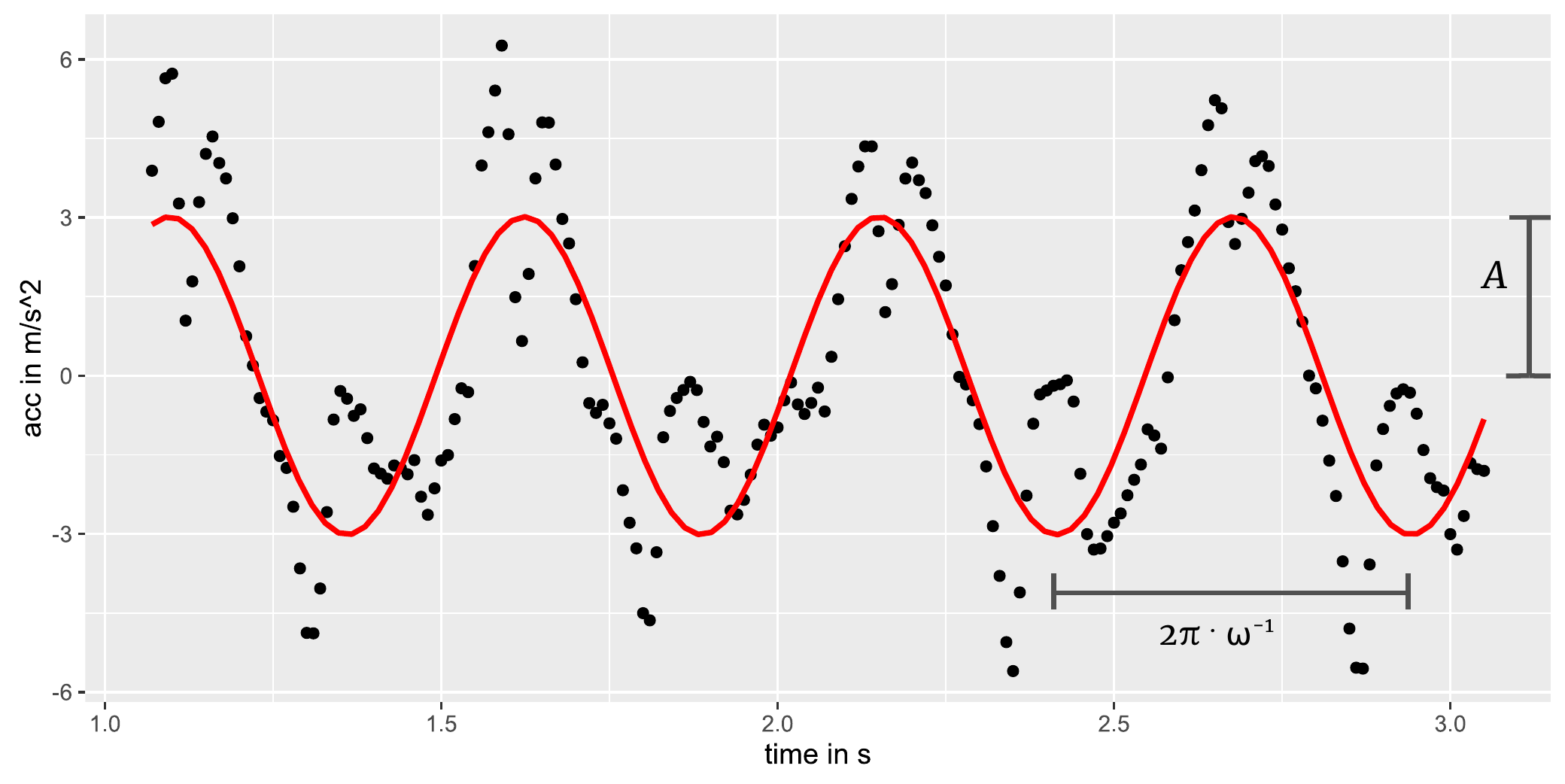}
    \caption{Fitting of the sinusoid model to the accelerometer data.}
    \label{fig:fitting}
\end{figure}

The approach is to fit the time series of the accelerometer readings to a sinusoidal function. This is possible because of the periodic nature of CPR (see Figure \ref{fig:fitting}). The sought CPR quality parameters can then be derived directly from the parameters of the sinusoidal function. The generic parameterized sine function can be written as follows:
\begin{equation}
    \hat{f}\left(t\right)=A\cdot sin\left(\omega t+\rho\right)+D
    \label{eq:sine1}
\end{equation}

Parameter $A$ and $\omega$ are of primary interest: $A$ is the amplitude, $\omega$ the angular frequency ($D$ is the vertical displacement, $\rho$ the phase shift). $A, \omega, \rho$ are adapted so that the sine curve fits the accelerometer data.
Assuming that the arms of a person performing CPR are orthogonal (and rigid) on the patient's chest, the relative movements of the bystanders' arms are equal to the chest compression depth. It is also assumed that the chest is wholly relieved after each compression (full chest recoil as recommended by ALS/BLS guidelines). Additionally, the frequency of low to high to low compression depth represents one compression cycle.
Since it is not applicable to fit the accelerometer data (total acceleration) directly to the sine curve (see Equation \ref{eq:sine1}) and derive the displacement of the arm from it, the accelerometer data must be integrated twice to determine the vertical displacement ($\int\int$ acceleration $\rightarrow \int$ velocity $\rightarrow$ displacement). However, the double integration of the acceleration values induces errors since the accelerometer can not be assumed to be perfectly calibrated, so this is rarely a practical way. To avoid this integration errors, we use the second derivative of the sine function as a model function. Thus, the analytical solution of the double integration is already known ($\hat{f}$, see Equation \ref{eq:sine1}):

\begin{equation}
    f\left(t\right)=\frac{\partial^2}{\partial t^2}(A\cdot sin\left(\omega t+\rho\right)+D)
\end{equation}

\begin{equation}
    f\left(t\right)=-A\omega^2\cdot sin\left(\omega t+\rho\right)
    \label{eq:sin_function}
\end{equation}

On a fitted function, parameter $\omega$ is the CPR frequency (CCF), and $2\cdot A$ is the compression depth (CCD). To fit the function $f\left(t\right)$, we minimize the root mean squared error (RMSE). Thus, we formulate the minimization problem (loss function $L$) as follows:

\begin{equation}
    L = \min\sqrt{\frac{1}{\left|a\right|}\sum_{t=0}^{T}\left(a\left(t\right)-y\left(t\right)\right)^2}
    \label{eq:minimization_problem}
\end{equation}

In Equation \ref{eq:minimization_problem}, $a$ is the T-dimensional tuple of accelerometer measurements (total acceleration or vertical acceleration), $\left|a\right|$ is the dimension of the tuple $a$, i.e., its number of elements. $T$ is derived from the window length $S_{len}$ (e.g., $S_{len}=3s$ with 100 Hz $\rightarrow T = 300$), which has been optimized in \cite{Lins2019}.

\subsection{Evolutionary Fitting of the Model Function}

In order to estimate the three variables $\omega, A, \rho$ of the sine function $f$ (Equation \ref{eq:sin_function}), herein a Evolution Strategy inspired algorithm is applied (see Algorithm \ref{alg:algorithm}). The algorithm uses also elements from GA, e.g. it is population-based and optimizes the population throughout several generations (($\mu+\lambda$)-Evolution-Strategy based on \cite{Simon2013} but with a GA-like population). 
Population-based algorithms can use the diversity of their population to optimize several local minima (in case of a minimization problem) in parallel over generations. In the case of the sine model, there are not several different local minima (only periodic ones), so only $\lambda$ individuals of the population are mutated and recombined per generation $G$. This reduces the computational effort per generation. The population-element of the algorithm is used as archive for possible good solutions.
The optimization is done between a transition from one generation $G$ to the next generation $G+1$. Within each transition from one generation to the next, the algorithm comprise the steps \emph{mutation}, \emph{crossover}, and \emph{selection}, which are discussed in the next sections. 
Algorithm \ref{alg:algorithm} is continuously applied on $S_{len}$ long data windows while retaining fractions of the population until $G = G_{max}$ or the convergence of $G$ is smaller than $c_{min}$.

\begin{algorithm}[h]
\SetAlgoLined
\KwData{Timeseries of accelerometer data with 100 Hz sampling rate}
\KwResult{A, $\omega$, $\rho$}

\While{sensor data available}{
{Every $f_U$, get next window $S$ with length $S_{len}$\;}
\eIf{first start}
{initialize population on first run\;}
{re-initialize fraction $\varepsilon$ of the population\;}
Evaluate all individuals, $x_0$ is currently best solution\;
\While{$G < G_{max} \texttt{ and }$ convergence $>= c_{min}$}{
Create completely random individual $\hat{x}_{\mu+1}$\;
Mutate $x_{0}$ to form  $\hat{x}_{\mu+2}$\;
Create $\lambda-2$ offspring $\hat{x}_{\mu+3}$ to $\hat{x}_{\mu+\lambda}$ using random chosen parents\;
Evaluate all individuals with cost function and window $S$\;
Remove $\lambda$ worst individuals from population\;
}
\textbf{output} model parameter A, $\omega$, $\rho$ of $x_{0}$
}
\caption{Pseudocode of used evolutionary algorithm.}
\label{alg:algorithm}
\end{algorithm}

\subsubsection{Initialization of the Parameter Space}

Be $x_{i,G}$ a 3-dimensional vector (containing sought parameter $A, \omega, \rho$) of individual $i$ for generation $G$:

\begin{equation}
    x_{i,G}\mathrm{\ with\ }i \in [1,\mu], G \in [1,G_{max}]
\end{equation}

In every generation $G$ $\lambda=1+1+3=5$ individuals are newly generated ($\hat{x}_{\mu+1}$ to $\hat{x}_{\mu+\lambda}$). Every of the $\mu$ individuals represents a possible solution to the minimization problem (see Equation~\ref{eq:minimization_problem}).

At the first start, the population is initialized by drawing parameters from an uniform random distribution (exploration). In the optimization process, individual solutions will move closer and closer to the optimum, and the variance within the population will decrease (exploitation).
Every new individual of the population is also initialized taking random variables from a uniform random distribution to foster the exploitation of the parameter space. For the specific task of fitting a sinusoidal function, the interval of the random distribution can be restricted to the problem's bounds. Table \ref{tab:bounds} summarizes the parameter space over all dimensions for the given task of fitting the CPR sinusoidal model.

\begin{table}[ht]
    \centering
        \caption{Parameter limits for every dimension used for initialization. The individuals are initialized by drawing random numbers from the given intervals using a uniform random distribution.}
    \label{tab:bounds}
    \begin{tabular}{c|c|c}
        Individual $x$  & Function Parameter Interval & CCF/CCD \\\hline
        $x_{i,G}(0)$ & $A \in [0.1, 5.0]$& CCD is $2A$ in cm\\
        $x_{i,G}(1)$ & $\omega \in [\pi, 7\pi]$  & CCF is $\frac{\omega}{2\pi} \cdot 60$ in cpm\\
        $x_{i,G}(2)$ & $\rho \in [0, 2\pi]$  & only used for model fitting \\
    \end{tabular}
\end{table}

\subsubsection{Mutation}
In the applied approach, mutation occurs in two ways. The first child $\hat{x}_{\mu+1}$ is simply a newly generated random individual. This ensures that the algorithm can still exploit the parameter space. Then, in each generation $G$, the currently best individual $x_0$ is mutated minimally and forms the second child $\hat{x}_{\mu+2}$ of the current generation. This mutation is a fine-tuning element to optimize the current solution.
The following $\lambda-2$ offspring are created as follows (with constant $M = 0.999$), where $\mathcal{U}$ generates a random value from a uniform random distribution within the interval: 

\begin{equation}
    m_k = \mathcal{U}(M, 2 - M)
\end{equation}

\begin{equation}
    \hat{x}_{\mu+2}(k) = m_k \cdot x_{0}(k) \text{ for } k \in [0,1,2]
\end{equation}

$k$ is the index of the 3-dimensional solution vector (see Table \ref{tab:bounds}).

\subsubsection{Crossover}
The crossover step is used to transfer information, i.e., possible solutions, from the current generation to the next generation and to recombine it. In the approach chosen here, three ($\lambda - 2$) offspring per generation are generated from two randomly selected parent individuals.

\begin{equation}
    x_a \text{ with } a = \mathcal{U}(1, \mu)
\end{equation}

\begin{equation}
    x_b \text{ with } b = \mathcal{U}(1, \mu), b \neq a
\end{equation}

\begin{equation}
    \hat{x}_{\mu+i}(k) = \mathcal{U}(min(x_a(k), x_b(k)), max(x_a(k), x_b(k))) \text{ with } k \in [0,1,2]
\end{equation}

This crossover method corresponds to the blended crossover (or BLX-$\alpha$ or heuristic crossover) with $\alpha = 0$ \cite{houckgenetic,herrera1998tackling}.

\subsubsection{Selection}

The selection step determines which individuals are passed to the next generation by evaluating them against the cost function. Herein, the RMSEs are summed up for every solution candidate $x_{i,G}$ as cost function:

\begin{equation}
    \sum_{t=0}^{T}\left(a\left(t\right)-f_{x_{i,G}}\left(t\right)\right)^2
\end{equation}

with $a$ being a $T$-dimensional tuple $S$ of samples (accelerometer data) and $f_{x_{i,G}}$ the parameterized sinusoidal function of individual $x_{i,G}$.
After the steps mutation and crossover $\mu+\lambda$ individuals exist. The population is therefore reduced to the $\mu$ fittest individuals after the evaluation of the population.
Once the optimization has finished, i.e. when $G_{max}$ or $c_{min}$ has been reached, the individual with the lowest RMSE represent the parameter of the sinusoidal model ($x_0$). From these parameters, the CCF and CCD can be obtained with $CCF=\frac{\omega}{2 \pi} \cdot 60$ cpm and $CCD=2A$ cm.

\subsection{Reconsideration of the Population}
Herein, the algorithm is initially only executed for a small window of 3 seconds ($S_{len} = 3$) of IMU data. Once a subsequent data-window is available (next 3s with 2s overlap $\rightarrow f_U = 1s^{-1}$), the algorithm is executed for a new fitting run. In a previous approach \cite{Lins2019}, the evolutionary optimization was performed with a completely new population. Possible solutions that are very similar to the current solutions must always be found and optimized anew. In contrast, herein a part of the population is retained and only one fraction $\epsilon$ is completely reinitialized.
On the one hand, a large diversity of the initial population can be ensured. On the other hand, good solutions from the past are retained in the expectation that they will also fit the current data window with a few adjustments. Fraction $\epsilon$ is being optimized in the hyperparameter optimization step (see \ref{sec:appendix}).

Initially, the favorable hyperparameters of the algorithm ($\mu, G_{max}, \epsilon, c_{min}$) are determined based on simulated data of sinusoidal curves (see \ref{sec:appendix}). With these specific hyperparameters, the algorithm is then evaluated within a study with human participants.

Based on the results of Monte Carlo simulations, we select $\mu = 400$, $G_{max} = 10$, $\epsilon = 0.5$, $c_{min} = 0.05$ as suitable hyperparameters for further evaluation with human participants.

\subsection{Experimental Setup}

The Laerdal Resusci Anne Simulator mannequin was used as a reference system and was placed on the floor (see Figure 3). Within the Resusci Anne simulator, sensors measure the depth of thorax compression and decompression, the frequency of the compressions, and the volume of ventilation.

\begin{figure}[ht]
    \centering
    \includegraphics[width=\textwidth]{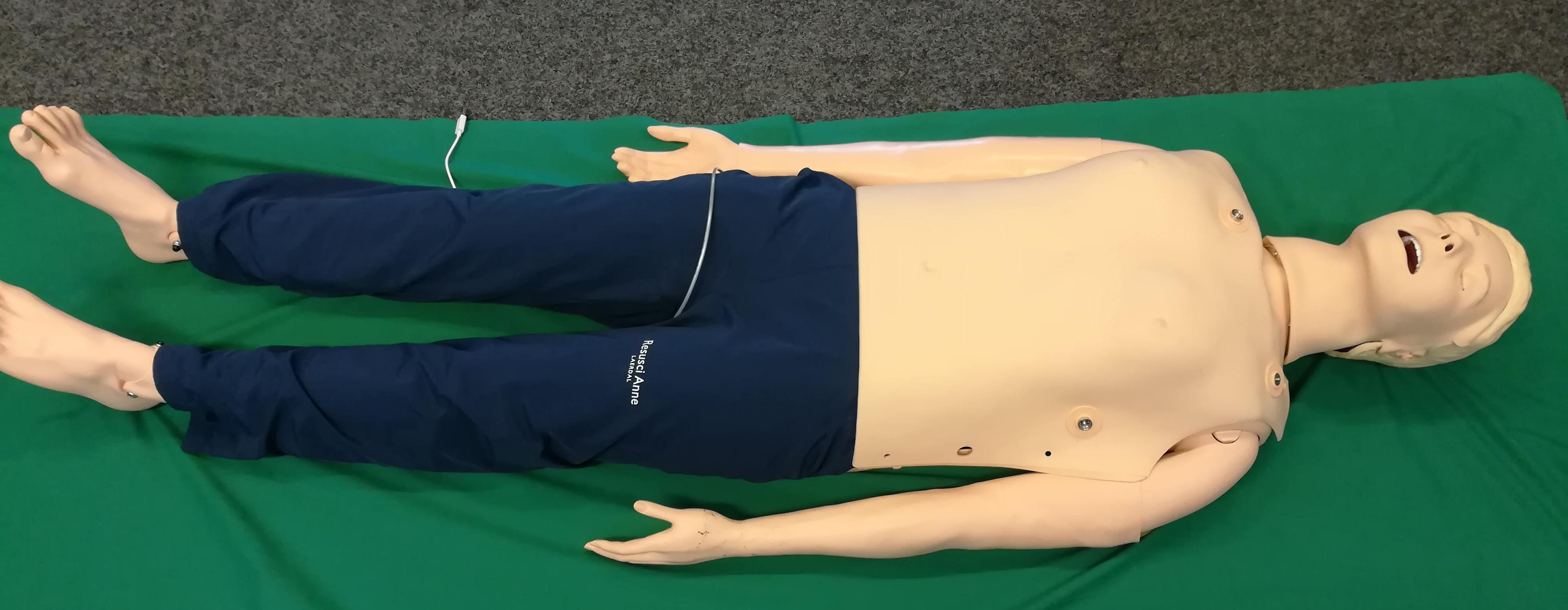}
    \caption{Resusci Anne training mannequin.}
    \label{fig:resusci_anne}
\end{figure}

\begin{figure}[ht]
    \centering
    \includegraphics[height=3cm]{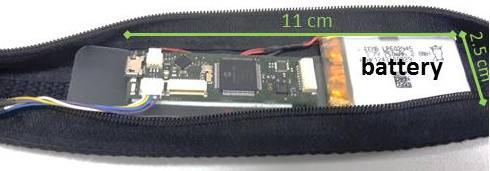}
    \includegraphics[height=3cm]{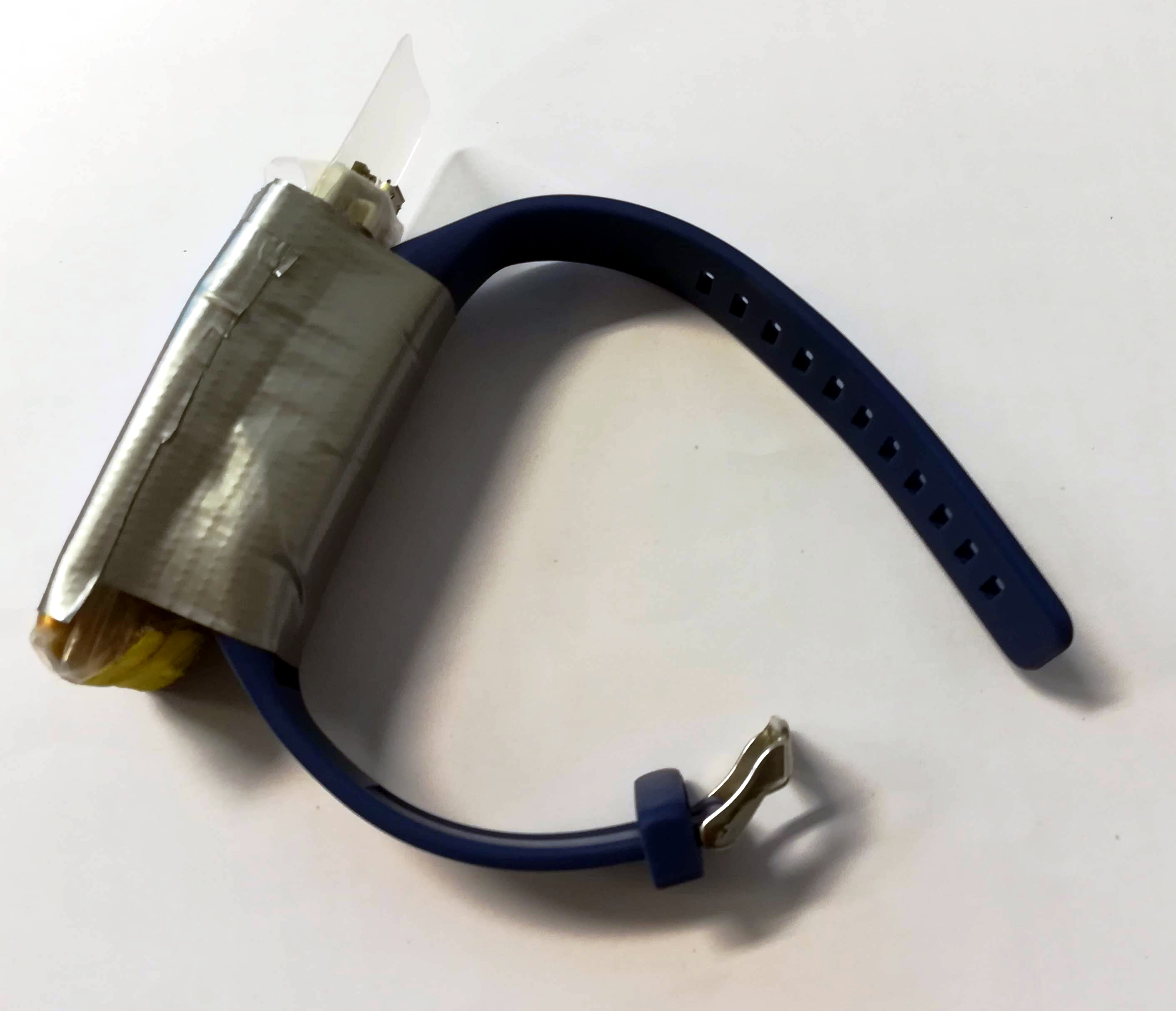}
    \caption{IMU sensor used in the study unpacked (left) and with bracelet (right).}
    \label{fig:sensor}
\end{figure}

While conducting CPR trainings on the Recusci Anne, the participants have been equipped with two IMU sensors (see Figure \ref{fig:sensor}) to evaluate the relevance of their placement. One IMU sensor was placed at the left wrist of the participant with a bracelet, and the other one was placed between the hands of the participant (between the back of the hand of the first hand and the palm of the second hand). Otherwise, no specific arm or upper body posture was explicitly required. The sensors and the training mannequin were collecting data that was synchronized manually after the recording via visual inspection of the accelerometer signal.
 
The 9DOF-IMU sensor contains a Bosch BMA180 triaxial accelerometer with sensitivity ranges from 1G up to 16G. A sampling rate of 100 Hz was selected, which is considered a typical sampling rate on smartwatches and also represents a reasonable compromise between computing effort and precision.

The participants were asked to place themselves on any side of the training mannequin and perform the CPR within 80-140 cpm.  The study director monitored the correct implementation of the CPR and, where necessary, corrected with verbal advice. The recording lasted for two minutes.

\subsection{Statistical analysis} 

\begin{figure}[hbt]
    \centering
    \includegraphics[width=0.7\textwidth]{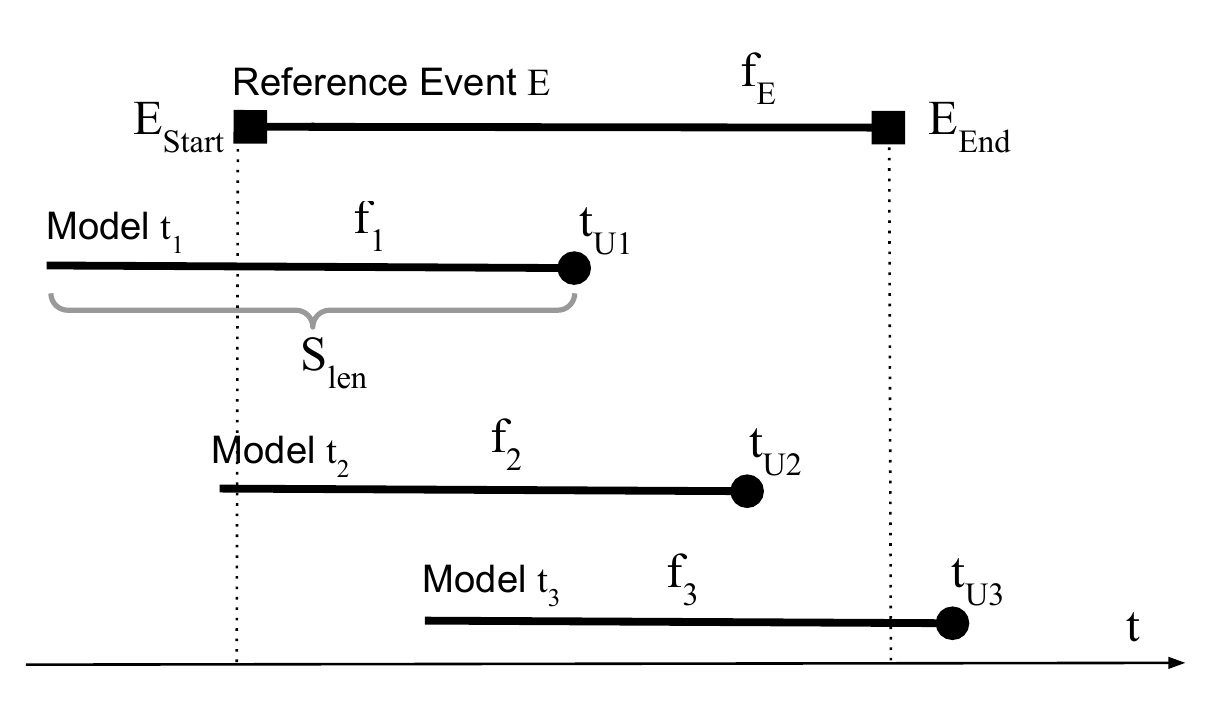}
    \caption{How model predictions and reference values are compared.}
    \label{fig:model_comparison}
\end{figure}

For every compression cycle recognized by the reference system, a 3-second window $S$ ($S_{Len} = 3s$) of the sensor data is used to fit the sinus model every second ($f_U = 1s^{-1}$) as recommended by Lins et al. \cite{Lins2019}.
Here, it is $S_{len} > f^{-1}_{U}$, so the used datasets $S$ are interleaved (see Figure \ref{fig:model_comparison}). 

Also, one Resusci Anne event $E$ (a complete compression/decompression cycle) may be smaller, equal, or larger than $f^{-1}_{U}$ so that we must combine one or more model predictions before comparing it with $E$ (Equation \ref{eq:pred_combine}).
Thereby, the weighted mean of $n$ subsequent model predictions within each interval $(E_{Start}, E_{End})$ is calculated with the overlap ratio $\sigma$ representing the weight:

\begin{equation}
\label{eq:pred_combine}
    p(t) = \frac{1}{\sum_{i=1}^{n} \sigma_i} \sum_{i=1}^{n} \sigma_i f_i
\end{equation}

For each compression event $E$, a corresponding prediction $p$ is determined according to Equation \ref{eq:pred_combine}. 

The measurement-error is calculated as the absolute difference between predicted CCF/CCD and the CCF/CCD of the reference system for every CCF/CCD prediction over every participant.

For comparison with the predictions of the sine models,  one-dimensional 1000-point Discrete Fourier Transformations (DCT) were calculated for the identical windows using the FFT algorithm.

\FloatBarrier
\section{Results}
 
The study cohort consisted of 18 participants, aged 19-48, 12 male, 6 female. The participants were recruited amongst students and staff of the University of Oldenburg, Germany. 
The study received ethics approval number “Drs.EK/2018/31” of the IRB of the University of Oldenburg.

\begin{table}[h]
    \centering
        \caption{Results summary: Absolute median [min, max] errors between predictions and reference values (lesser is better).}
    \begin{tabular}{c|c|c}
        Position   &  Hand &  Wrist \\\hline\hline
        CCF (Sine Model) & 2.38 [0.0-108.7] cpm & 2.22 [0.0-110.8] cpm\\
        CCF (FFT) &  8.05 [0.0-127.9] cpm & 7.42 [0.0-120.0] cpm \\\hline
        CCD (Sine Model) & 0.64 [0.0-3.89] cm & 0.69 [0.0-5.03] cm 
    \end{tabular}

    \label{tab:result_overview}
\end{table}

Table \ref{tab:result_overview} contains the overall results of the comparison between model predictions and reference data. The results show the least error ($\pm 2.22$~cpm) for CCF prediction for the sinusoidal model based on the sensor data from the wrist-worn sensor. The alternative method FFT shows a much higher error with $\pm 7.42$~cpm. When predicting the compression depth CCD, the wrist sensor's data with $\pm 0.69$~cm show an slightly higher error as the hand-held sensor, which median error is $\pm 0.64 $~cm.

\begin{figure}
    \centering
    \includegraphics[width=\textwidth]{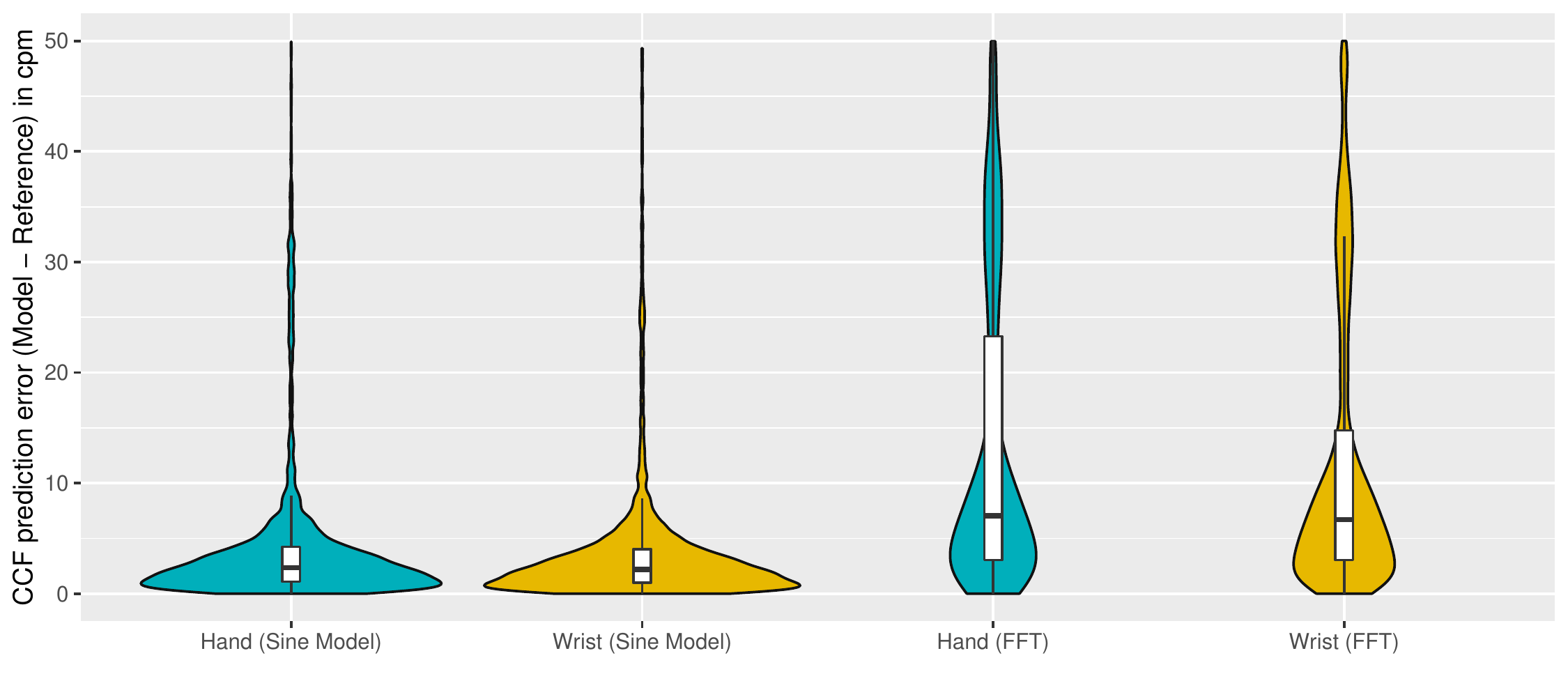}
    \caption{Absolut error distribution of models/methods for CCF prediction (range [-50, 50]). }
    \label{fig:result_ccf_pred}
\end{figure}

Figure \ref{fig:result_ccf_pred} shows the error distribution of the CCF prediction of the sine model versus FFT considering the two sensor positions. The sine models' errors are predominantly in the area of $[0, 5]$ cpm showing little outliers.
The error distribution for the FFT variants is mainly in the area of about $[0, 15]$ showing more outlier sets between $[25, 50]$ cpm. 

\begin{figure}
    \centering
    \includegraphics[width=\textwidth]{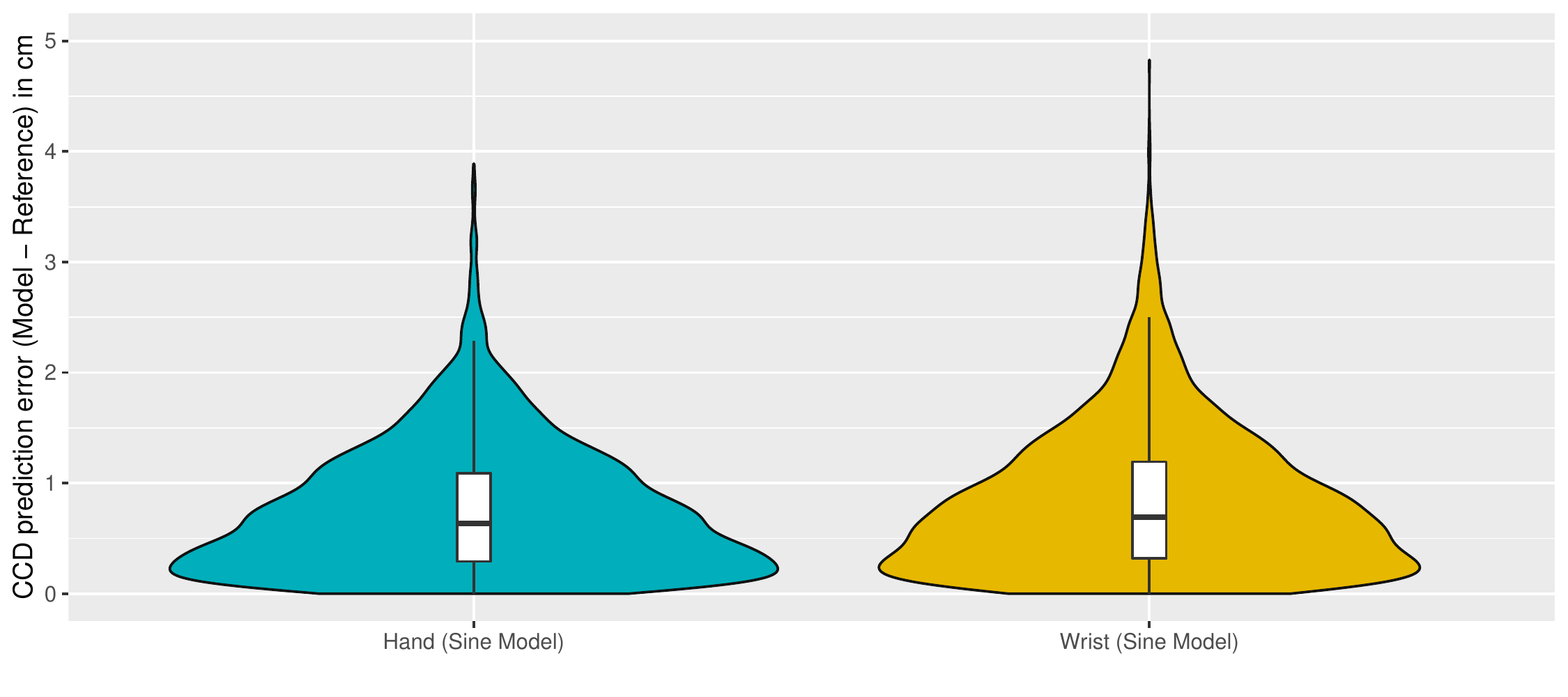}
    \caption{Absolute error distribution of models/methods for CCD prediction.}
    \label{fig:result_ccd_pred}
\end{figure}

Figure \ref{fig:result_ccd_pred} shows the error distribution of the CCD prediction of the sine models for the two considered sensor positions. The distribution of the model based on the hand-held sensor is centered around $0.5$~cm with no notable outlier groups (some outliers up to $4$~cm). The wrist-worn sensor's distribution is also centered around $0.5$~cm, also with no notable outlier groups but some outliers up to $5$~cm.

\begin{figure}[h]
    \centering
    \includegraphics[width=0.79\textwidth]{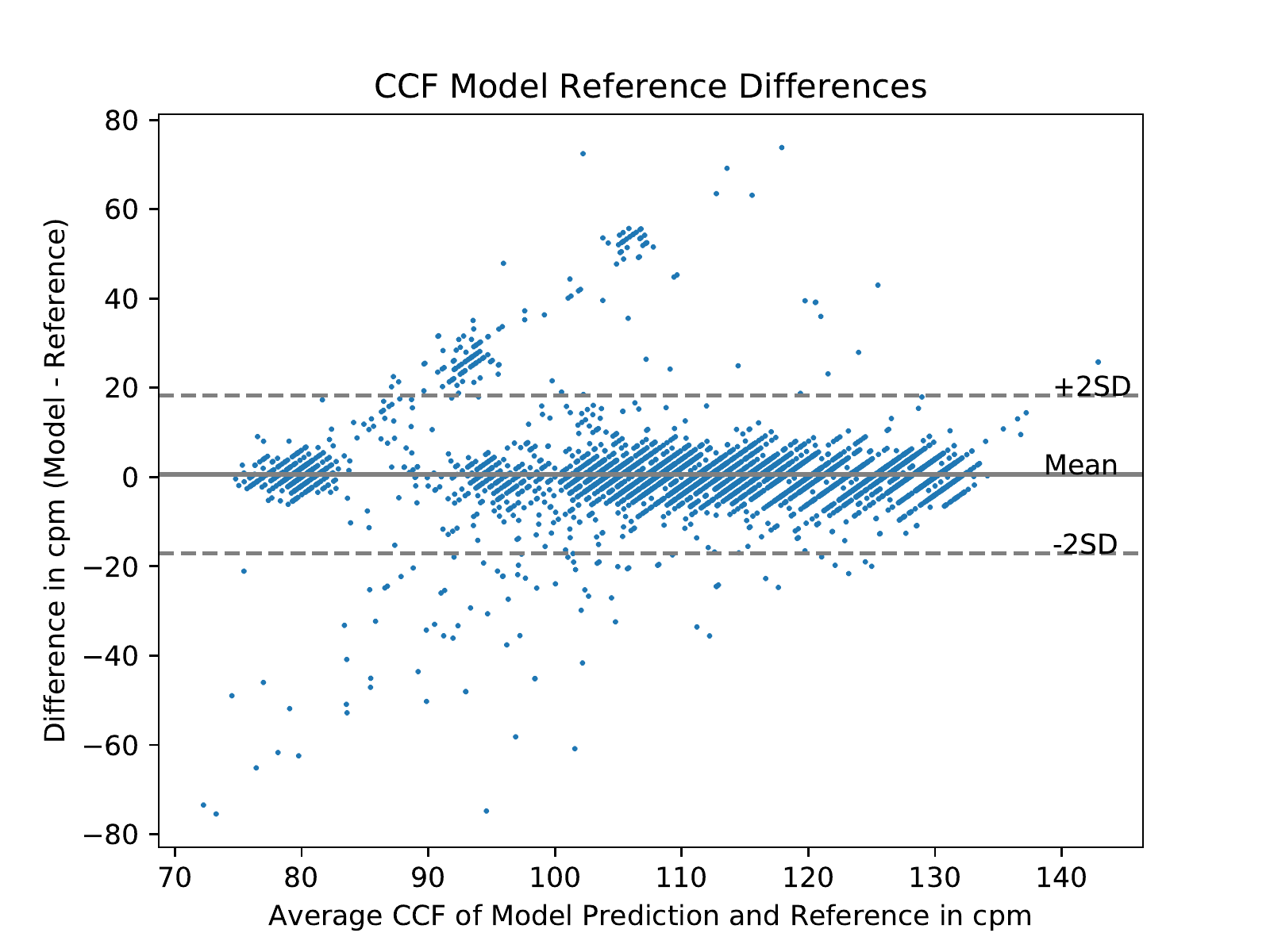}
    \caption{Differences against mean CCF measurements using wrist-worn sensor.}
    \label{fig:bland_altman_ccf}
\end{figure}

\begin{figure}[h]
    \centering
    \includegraphics[width=0.79\textwidth]{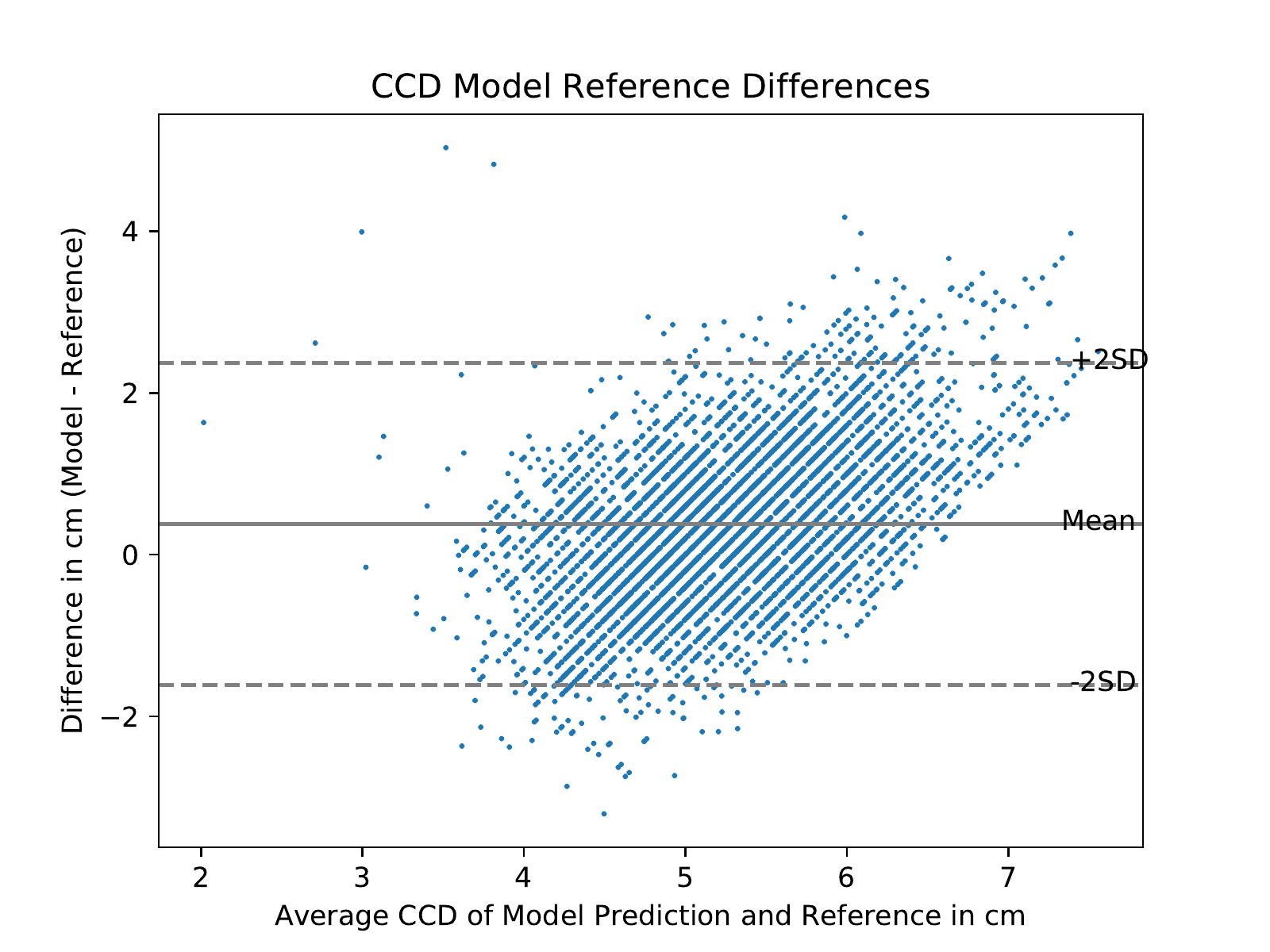}
    \caption{Differences against mean CCD measurements using wrist-worn sensor.}
    \label{fig:bland_altman_ccd}
\end{figure}

In Figures \ref{fig:bland_altman_ccf} and \ref{fig:bland_altman_ccd} -- unique forms of point diagrams (Bland-Altman plots) -- the differences between the predictions of the sinusoidal models and the measurement of the reference system are plotted against the mean value of the two methods. 
For the compression frequency (Figure \ref{fig:bland_altman_ccf}) the error is within one standard deviation in most cases. A closer look reveals a linear relationship between increasing error and measured frequency (diagonal arrangement of data points). 
In case of the compression depth (Figure \ref{fig:bland_altman_ccd}) most measurements and predictions are between 4-6.5~cm and within two standard deviations and few outliers.

\FloatBarrier
\normalsize
\section{Discussion}

In this study, two research questions have been investigated.
To clarify the suitability of smartwatch-like inertial sensors for the detection of the chest compression frequency (CCF) and depth (CCD) during CPR in comparison to the typical hand-held smartphone sensor, both sensor positions were evaluated with the Resusci Anne training mannequin as reference system.

With an error of $\pm2.22$ cpm, the CCF accuracy of the wrist sensor was slightly more accurate  compared to an error of $\pm2.38$ cpm for the common grasp-in-hand use. Both errors remain in the same order of magnitude, confirming the algorithm's robustness and the general suitability of the sinusoidal model-based approach. Consequently, we could confirm the suitability of the smartwatch-related wrist positioning for CCF calculation.

Considering the CCD, we found a slightly higher error (0.69 cm) for the wrist sensor when compared to the grasp-in-hand use (0.64 cm). Nevertheless, even with the relatively high error of the wrist sensor, a fundamental qualitative statement can be made about the compression depth, since relevant deviations from the optimal compression depth of 5-6 cm (e.g., too low compression depths of 3 cm) might be still detectable. However, it should be noted that the CCD prediction errors are probably too high for quantitative statements.

Concerning the second research question, the applicability of the evolutionary approach for curve-fitting of the chest compression during CPR was examined, especially regarding the accuracy of resulting CCF and CCD parameters for the considered two sensor placements in comparison to the Resusci Anne as a reference system.
The predictions of the sinusoidal model were compared to the FFT method and achieved significantly low error values both for sensor position hand (Sine Model: $\pm2.38$~cpm, FFT: $\pm8.05$~cpm) and sensor position wrist (Sine Model: $\pm2.22$~cpm, FFT: $\pm7.42$~cpm).
With an error of $\pm2.22$ cpm in predicting the CCF, it is higher than the range given in the literature. Even though Ruiz de Gauna et al. \cite{DeGauna2016} reported a lower median error of 0.9 cpm for their approach, differences among both studies could as well be affected by variations among reference system (photoelectric sensor) and IMU sensor (ADXL330, Analog Devices, USA). It thereby might not only be related to the algorithmic specifics. 
In any case, the error of 2.22 cpm is far below the critical target range of 100-120~cpm and thereby does not affect the practical applicability of the algorithm.

In comparison to the related approaches, the proposed system achieves a comparable CCF accuracy within the ERC guideline requirements of $\pm10$ cpm \cite{Perkins2015}. However, we found a high inter-subject variability of the sensitivity to measure the CCD. 

Consequently, wrist-worn devices can accurately predict the CCF, while CCD measure can be expected to hold an error of up to 0.69~cm. Although CCD prediction still shows potential for improvement, the results are sufficient for practical applications in smartwatches to make qualitative statements about the quality of the CPR. Smartwatches are therefor a well suited, unobtrusive, and high available alternative platform for giving CPR feedback for bystanders in emergencies. 

While the proposed CPR modeling via a sinusoidal model is intended for an online system, it was tested as an offline system for evaluation purposes. Due to its iterative nature, the sine model fitting can probably be used well on (mobile) embedded devices. In contrast, the population-based optimization of the evolutionary algorithm is well suited for parallelization on multi-core systems, which become defacto-standard in current mobile devices. 
As the algorithm can be dynamically adapted to varying processing power conditions, e.g. by adjusting hyperparameters such as number of generations and individuals the given approach is well suited for use in smartwatch application where even with cheaper hardware good results can be achieved, albeit not with the highest precision. The convergence ensures that each offspring generation is at least as good as the previous one, and even in limited generations, models can be expected to be fine-tuned to the physiological characteristics of the users and the sensor environment.

A more complex sine function, such as a linear combination of several sine terms, may be also an appropriate model function. For the usage scenario here, namely to determine the frequency and compression depth of the CPR in a straightforward way, a more complex function would be a hindrance, even though it might fit better to the data. Therefore a comparatively simple sine function was chosen here.

\FloatBarrier

\section{Conclusion}

The article introduces and evaluates an approach to predict frequency and depth parameters of CPR training via accelerometer data of a wrist-worn IMU in comparison to a hand-held IMU. The sinusoidal models for estimating the compression frequency and depth were dynamically adapted using a Evolution Strategy inspired evolutionary algorithm.
The approach was evaluated with 18 human participants for both hand-held and wrist-worn sensor positions. The chest compression frequency (CCF) was predicted with a median error of $\pm2.22$ cpm and the compression depth (CCD) with a median error of $\pm0.69$ cm for the wrist sensor position. 
Thus, our work represents an essential step towards complete and precise modeling of the CPR using mobile sensors as can be found in smartwatches. While focusing on the algorithmic aspects of the detection of the CPR parameters, the general feasibility of smartwatches for CPR feedback (e.g. via a full-featured smartwatch application) has to be investigated further. The next steps are the development of a corresponding smartwatch application and subsequent usability studies.

\section*{Acknowledgements}
This work was supported by the funding initiative Nieders\"{a}chsisches Vorab of the Volkswagen Foundation and the Ministry of Science and Culture of Lower Saxony as a part of the Interdisciplinary Research Centre on Critical Systems Engineering for Socio-Technical Systems II.

\section*{References}

\bibliographystyle{elsarticle-num.bst}
\bibliography{cpu_imu}

\appendix

\section{Evaluation with simulated data}
\label{sec:appendix}

For the simulation-based optimization, periodic acceleration data for ten random frequencies were generated within the interval [1, 3] Hz (60-180 cpm). For each frequency, a three-second sample was generated using the model function (see Equation \ref{eq:sin_function}) and normal-distributed Gaussian noise was applied. The amplitude of the function was randomly selected from [0.01, 0.05] m. 
This one test round with ten random frequencies was considered as one individual for a meta evolution using the Differential Evolution (DE) algorithm (Python/scipy implementation).

A cost function was chosen that considers the number of individuals and the maximum number of generations (less is better):

\begin{equation}
    cost = \mu \cdot G_{max} \cdot \sum_{i=1}^{10} |M_{CCF} - F| \cdot 
        \sum_{i=1}^{10} |M_{CCD} - D| 
\end{equation}

with $M$ being the number of model predictions and $F$ and $D$ the random frequency and depth values. The DE algorithm was optimizing with a population size of 100 and 100 iterations.

\paragraph{Simulation Data Results}
The meta-optimization of the hyper parameters using the DE algorithm produced the near optimal values shown in Table \ref{tab:results_meta}. Since the results of heuristic algorithms are subject to random fluctuations, the meta-optimization was performed 100 times in total and the deviations were noted. Table \ref{tab:results_meta} gives mean and median deviations of the optimizations.

\begin{table}[hb]
\small
    \centering
        \caption{Results of the hyperparameter optimization, Monte-Carlo-Simulations $N=100$}
    \label{tab:results_meta}
    \begin{tabular}{r|c|c|c|c|c}
        \textbf{Parameter} & $\mu$ & $G_{max}$ & $\epsilon$ & $c_{min}$ \\\hline
        \textbf{Mean [Min-Max]} & 404 [39-956] & 8 [5-24] & 0.5 [0.0-1.0] & 0.05 [0.0-0.1]\\
        \textbf{Median [SD]} & 334 [242] & 7 [3] & 0.5 [0.28] & 0.5 [0.03] 
    \end{tabular}

\end{table}

\end{document}